\documentclass[11pt,a4paper]{article}
\usepackage[hyperref]{emnlp2018}
\usepackage{times}
\usepackage{latexsym}
\usepackage{graphicx}
\usepackage{listings}
\usepackage{url}
\usepackage{dblfloatfix}
\usepackage{caption}
\usepackage{balance}

\aclfinalcopy 

\title{Discourse-Wizard: Discovering Deep Discourse Structure in your Conversation with RNNs}

\author{Chandrakant Bothe, Sven Magg, Cornelius Weber and Stefan Wermter \\
  Knowledge Technology, Department of Informatics \\
  University of Hamburg \\
  Vogt-Koelln-Str. 30, 22527 Hamburg, Germany \\
 \url{www.informatik.uni-hamburg.de/WTM/} \\
  {\tt \{bothe,magg,weber,wermter\}@informatik.uni-hamburg.de} 
}

\date{25.05.2018}

\begin{document}
\maketitle
\begin{abstract}
  Spoken language understanding is one of the key factors in a dialogue system, and a context in a conversation plays an important role to understand the current utterance. 
  In this work, we demonstrate the importance of context within the dialogue for neural network models through an online web interface live demo. 
  We developed two different neural models: a model that does not use context and a context-based model. 
  The no-context model classifies dialogue acts at an utterance-level whereas the context-based model takes some preceding utterances into account.
  We make these trained neural models available as a live demo called Discourse-Wizard using a modular server architecture. 
  The live demo provides an easy to use interface for conversational analysis and for discovering deep discourse structures in a conversation.  
\end{abstract}

\section{Introduction}

In recent research, spoken language understanding has received considerable attention due to its importance in dialogue systems. 
Discourse/conversational analysis can be performed by recognizing dialogue acts. 
The dialogue act defines the performative function of an utterance \cite{austin1962things,searle1979,allenCore1997}, and is also referred to as speech act in general. 
For many years, dialogue act (DA) recognition has been seen as an utterance-level classification \cite{stolcke2000dialogue,grau2004dialogue}.
However, a DA is a context-sensitive discourse concept \cite{grosz1982discourse,sbisa2002speech} and while collecting data, context sensitivity has been taken into account many times during the human (manual) annotation process, for example, for the Switchboard Dialogue Act (SwDA) corpus \cite{godfrey1992switchboard,danjur1997swbddamsl,Jurafsky1998}. 
SwDA is annotated with the Dialogue Act Markup in Several Layers (DAMSL) tag set \cite{allenCore1997} and the annotation of the current utterance is being performed by looking also at the preceding utterances. 
For example, the utterance \textit{'Yeah.'} is annotated as a \textit{\textbf{Yes-Answer}} type if it appeared after a \textit{\textbf{Yes-No-Question}} type of the DA. 
However, the same utterance was annotated as \textit{\textbf{Backchannel}} or \textit{\textbf{Accept/Agree}} type if it appeared after a \textit{\textbf{Statement}} type of the DA. 
Hence, in recent years, dialogue act recognition has been modeled using context-based approaches \cite{kalchbrenner2013recurrent,kumar2017dasl,ortega2017neural,Meng2017}.

\begin{figure*}[b!]
\begin{center}
\includegraphics[scale=1.0]{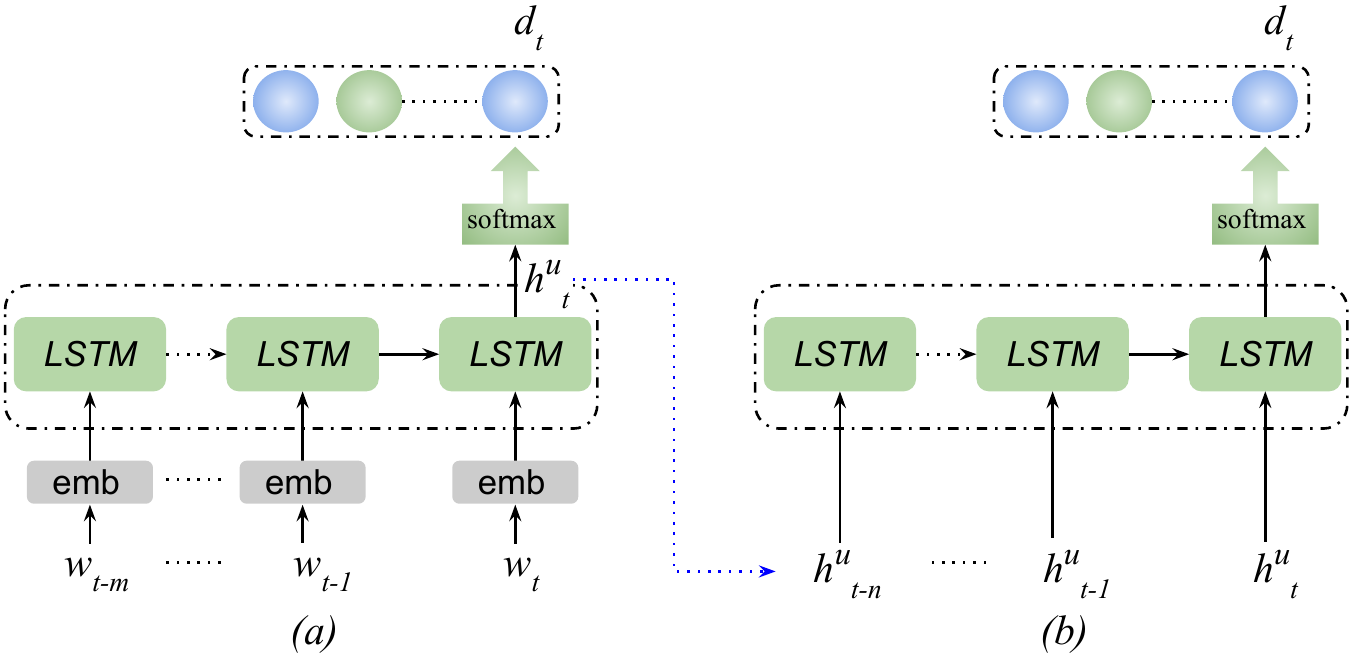} 
\caption{ (a) No-context model, (b) Context-based model.}
\label{non_contx_sent_rep}
\end{center}
\end{figure*}

In this paper, we introduce Discourse-Wizard tool to get your spoken (or written) conversation and analyze it for discovering its deep discourse structures with the help of dialogue acts.
We propose to use two models for the demonstration, a no-context model which performs an utterance-level DA classification and a context-based model which uses preceding utterances for learning the DA of the current utterance.

We host models on a server and developed an interactive web interface for discourse analysis. 
A user can input a set of utterances (a dialogue) with one turn/speaker utterance per line and get the DAs recognized from both models. 
The Discourse-Wizard Demo\footnote{\url{https://secure-robots.eu/fellows/bothe/discourse-wizard-demo/}} is available at the website of the EU SECURE (Safety Enables Cooperation in Uncertain Robotic Environments) Project.

\section{Related work}

Conversational analysis can be performed by analyzing the utterances for particular tasks like dialogue act recognition \cite{stolcke2000dialogue,grau2004dialogue}. 
However, utterances within a conversation are context-sensitive, and as most of the time, the dialogue act of the current utterance is based on the preceding utterances \cite{grosz1982discourse,sbisa2002speech}. 
Hence, modelling context-based approaches in conversational analysis becomes crucial \cite{kalchbrenner2013recurrent,ltAl2017EMNLP,ortega2017neural,Meng2017,BOTHE18_525}. 

On the other hand, there are many live demos available in the field of natural language processing for different tasks,
named entity detection, text tokenization,  part of speech tagging, sentiment analysis, and word embedding demos \cite{Loper2002NLTK:Toolkit,socher2013recursive,manning2014stanford,kutuzov2017building}.
In our work, we add discourse analysis to this list of useful demonstrations. 

\section{Approach}

In the following sections, we describe the two models used for the demonstration, the no-context model and the context-based model.

\subsection{No-context model}

The no-context model is a single utterance classification model.
A special recurrent neural network (RNN) model called long short-term memory (LSTM) is used to classify the dialogue act of the utterance as a sequence of word embeddings \cite{elman1990finding,wermter1995hybrid,Hochreiter1997LongMemory,lai2015recurrent}. 
LSTM-RNNs are chosen because of their advantage in sequential input modelling.
The model architecture is shown in Figure \ref{non_contx_sent_rep}(a), where the word one-hot vectors ($w_{t}, w_{t-1},...w_{t-m}$) are randomly initialized with vector representations called word embeddings. 
These word embeddings are learned during the training process with a multi-layer perceptron layer $emb$ through back-propagation. 

The LSTM learns hidden representation $h_{t}^{u}$ using $m$ number of words, 
at time step $t$ it is calculated as:
\begin{equation}
h_{t}^{u} = LSTM \left (w_{t}, w_{t-1},...w_{t-m}, \theta \right )  
\end{equation} 
where $\theta$ represents the hyper-parameters of LSTM such as embeddings, weight matrices and bias vector those are learned during the training process. 
We use a 50-dimensional embedding size and 64 hidden units for the RNN, these values were identified empirically.
The output $d_{t}$ is the dialogue act label of the utterance $u_t$ calculated using $h_{t}^{u}$, as:
\begin{equation}
d_{t} = g \left ( W_{out} \cdot h_{t}^{u} \right ) 
\end{equation}
where $W_{out}$ is the output weight matrix. 
All the hyper-parameters are further adapted using back-propagation through time. 
The task is to classify several classes; hence we use a $softmax$ function $g()$ at the output layer and categorical cross-entropy as a cost function.


\subsection{Context-based model}

The context-based model takes into account the preceding utterances while modelling the dialogue act of the current utterance. 
This is performed using hierarchical RNN (HRNN) layers \cite{el1996hierarchical,lee2016sequential,chung2017hierarchical}. 
The overall architecture is shown in Figure \ref{non_contx_sent_rep}(b), which takes input from already trained hidden layer of the no-context model. 

The utterances ($u_{t}, u_{t-1},...u_{t-n}$) are processed such that $u_{t}$ is the current utterance and there are $n$ utterances are in the context. 
Each utterance is passed through the no-context model and the hidden representations are used as an input to the second layer. 
Then last hidden vectors from the first layer are used as the utterance representations ($h_{t}^{u}, h_{t-1}^{u},...h_{t-n}^{u}$). The second-layer hidden vector ($h_{t}^{s}$) is calculated as: 
\begin{equation}
h_{t}^{s}= LSTM \left ( h_{t}^{u}, h_{t-1}^{u},...h_{t-n}^{u}, \theta \right )  
\end{equation} 
where $\theta$ represents the hyper-parameters of the network. The context-based model is trained similarly to the previous model to learn dialogue act ($d_{t}$) recognition while using the new hidden vector ($h_{t}^{s}$):
\begin{equation}
d_{t} = g \left ( W_{out} \cdot h_{t}^{s} \right ) 
\end{equation}
where the output weight matrix ($W_{out}$) is learned using back-propagation through time. 
Again, the $softmax$ function is used at the output layer. 
As a result, the $d_{t}$ in the context-based model is derived from the input utterances using hierarchical recurrent neural network:
\begin{equation}
d_{t} = HRNN \left (u_{t}, u_{t-1},...u_{t-n} \right ) 
\end{equation}
where $n$ is the number of utterances in the context.

\begin{table}[t]
\centering
\begin{tabular}{lll}
\hline
                         & train   & test   \\ \hline 
Number of conversations  & 1,115   & 19     \\ 
Number of utterances     & 196,258 & 4,186  \\ \hline
\end{tabular}
\caption{SwDA corpus details.}
\label{table:dataset}
\end{table}

\section{Experiments and results}

The models were trained on the Switchboard Dialogue Act (SwDA\footnote{Available at: \url{https://github.com/cgpotts/swda}}) corpus \cite{godfrey1992switchboard,danjur1997swbddamsl}. 
We have used the same dataset split as in \citet{stolcke2000dialogue} and \citet{kalchbrenner2013recurrent}, see Table \ref{table:dataset}.
The model was trained with the categorical cross-entropy loss function and using the \textit{softmax} function to classify the 42 dialogue acts ($d_{t}$) from the SwDA corpus. 

The results of the model performance are reported in Table \ref{table:Results}. 
The accuracy measure was used to compare the performance with the state-of-the-art results. 
The no-context model achieved 71.76\% of accuracy, which shows that the model surpasses the baseline performance. 
However, by applying the context-based model, the performance rose by about 2\% of accuracy with one utterance in the context ($n=1$). 
Adding utterances in the context improves the performance, but from the results, it seems that two utterances ($n=2$) are sufficient to capture information from the context, see also \citet{bothe2018conversational,BOTHE18_525}.
Hence, for the web-demo, we used the context-based model with $n=2$.

\begin{table}[t]
\centering
\begin{tabular}{llll}
\hline
Model setup                          &  Acc.(\%)   \\
\hline
\textit{Baseline and related work}   &           \\
Most common class                    &  31.50    \\

\citet{stolcke2000dialogue}          &  71.00    \\
\citet{kalchbrenner2013recurrent}    &  73.90    \\
\citet{lee2016sequential}            &  73.10    \\
\citet{ortega2017neural}             &  73.80    \\

\textit{Our work}                    &          \\
Non-utterance-context model          &  71.76   \\
Context-based model ($n=1$ utt.)    &  73.78   \\
Context-based model ($n=2$ utts.)   &  74.37   \\
Context-based model ($n=3$ utts.)   &  74.30   \\
Context-based model ($n=4$ utts.)   &  74.38   \\ 
\hline
\end{tabular}
\caption{Accuracy of the dialogue act detection. }
\label{table:Results}
\end{table}

\section{Web-demo}

\begin{figure}[b]
\begin{center}
\includegraphics[scale=0.36]{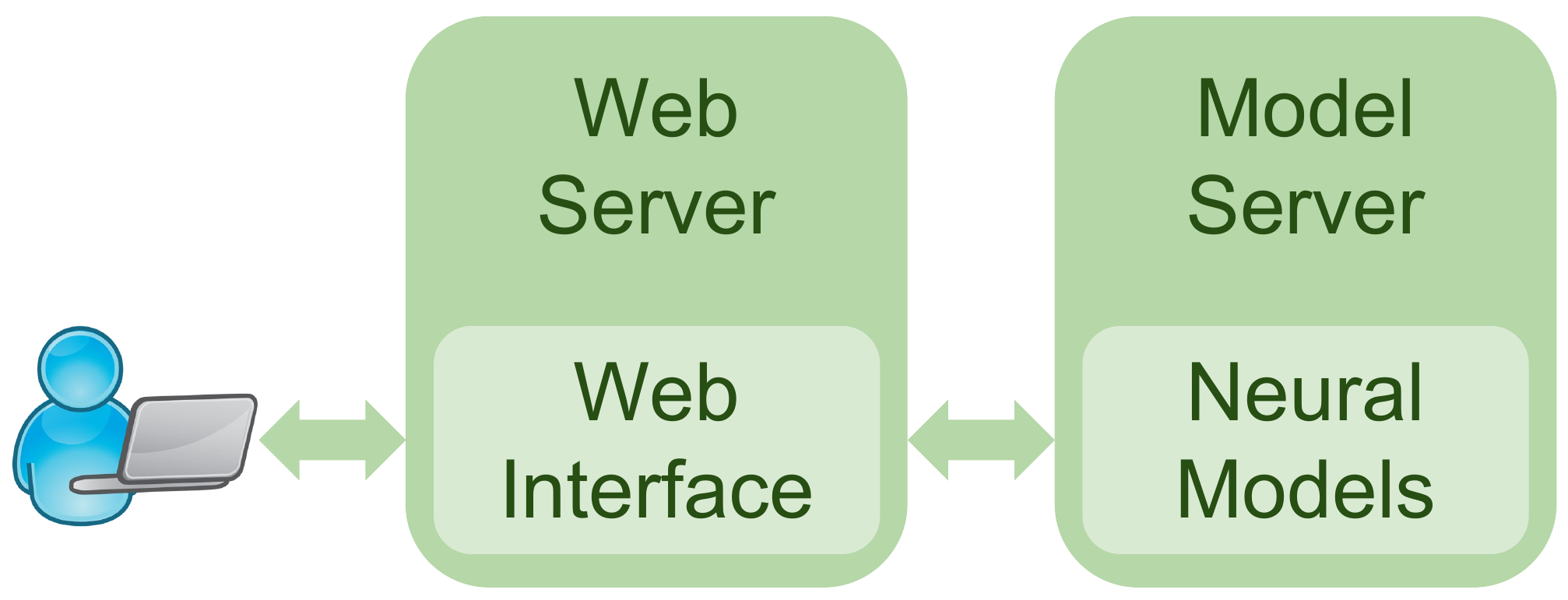} 
\caption{Architecture of the Web-demo}
\label{web-demo}
\end{center}
\end{figure}

The trained models are made available for the web-demo using a client-server architecture. 
The overall architecture is shown in Figure \ref{web-demo}. 

\subsection{Technical details}
The neural models are developed using Keras \cite{Chollet2015Keras} and TensorFlow \cite{abadi2016tensorflow}. 
Both models share similar properties and parameters, such as the vocabulary, embeddings, and most importantly, the context-based model uses the representation from the no-context model to encode the utterances.
Hence, it is possible to efficiently encapsulate these neural models within a Model Server.

\begin{figure*}[t]
\begin{center}
\includegraphics[scale=0.65]{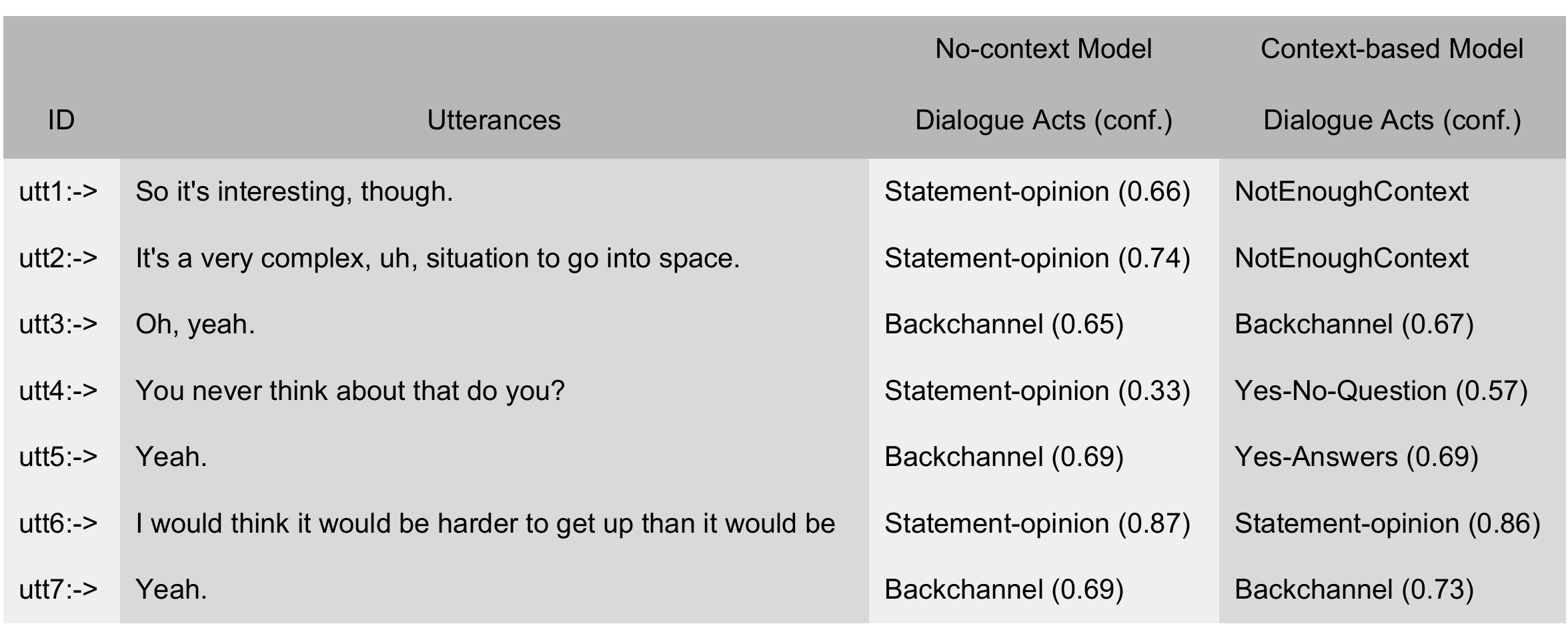} 
\caption{Example of output table generated from the web-interface. The input is only the set of utterances entered in the text box line by line. Each alternating line corresponding to one speaker.}
\label{FigExampleOfDemo}
\end{center}
\end{figure*}

This Model Server is developed using Flask, a micro web framework written in Python \cite{grinberg2018flask}. 
For the Web Server, we used the Django web framework \cite{holovaty2009definitive}, as it provides a template scheme for the web interface. 
We used two separate servers for hosting the models and for the web interface, for easier deployment and modular development.


The text input as a list of utterances from the Web Interface is sent as a request from the Web Server to the Model Server as shown in Figure \ref{web-demo}. 
The Model Server sends back the results as lists of recognized dialogues acts and confidence values (conf.).
The Web Server produces an output result as a summary and contains also a click-to-expand and -collapse table form as detailed results. 

\subsection{Analysis}

The output result produced on the Web Interface from the models is shown in Figure \ref{FigExampleOfDemo} for an example (cleaned) taken from the SwDA test set. 
The context-based model uses two utterances as context, therefore we cannot get the output for the first two utterances in the given dialogue, and the output is represented by \textit{NotEnoughContext}.
Notice for the utterances \textit{utt4} and \textit{utt5} that the no-context model failed to recognize the correct dialogue act class.
However, the context-based model could correctly recognize the dialogue act. 

Also, notice that the utterances \textit{utt6} and \textit{utt7} are the same in terms of syntax (the utterance \textit{"Yeah."}).
The no-context model classified both of these utterances as \textit{\textbf{Backchannel}} dialogue act. 
On the other hand, the context-based model could correctly learn that they belong to different classes because of the different contexts.
The same utterance is labeled as \textit{\textbf{Yes-Answer}} after the \textit{\textbf{Yes-No-Question}} and after the \textit{\textbf{Statement-opinion}} it is labeled \textit{\textbf{Backchannel}}, with quite high confidence.

\begin{figure*}[t]
\begin{center}
\includegraphics[height=6cm,width=16cm]{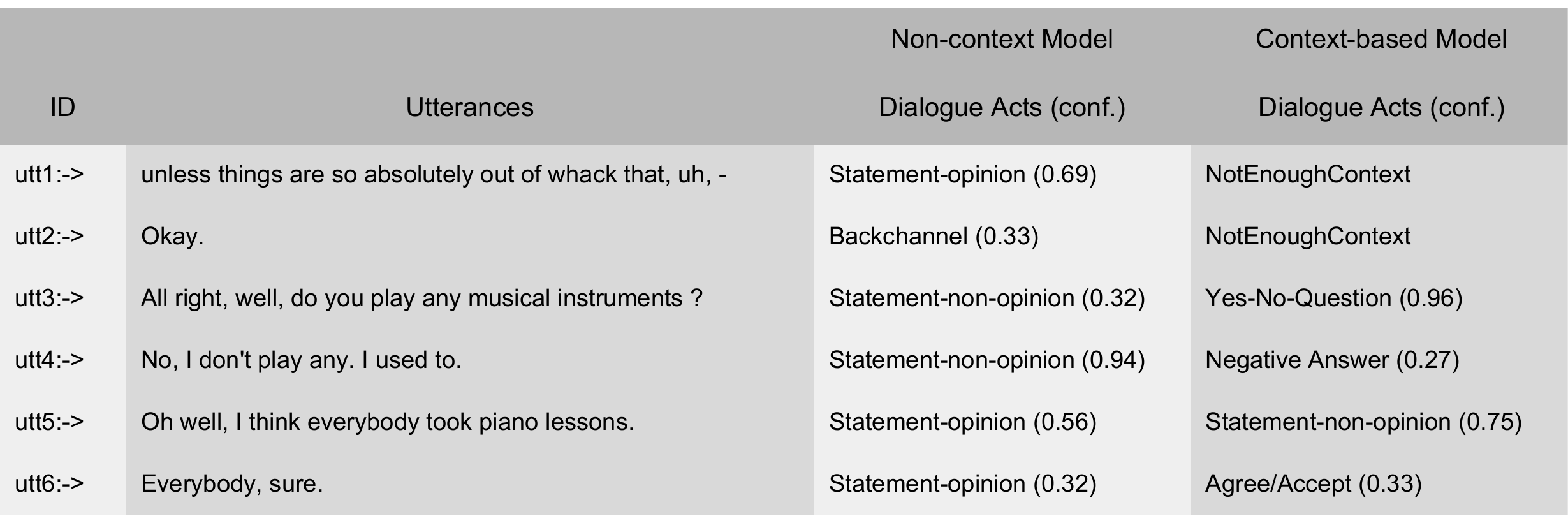} 
\caption{Some more examples of output generated from the web-interface. Particularly interesting to notice difference in the recognition of the utterances \textit{utt3}, \textit{utt4}, and \textit{utt6}.}
\label{FigExampleOfDemo1}
\end{center}
\end{figure*}

A few extra examples are provided in Figure \ref{FigExampleOfDemo1}, the \textit{utt3} is predicted as \textit{\textbf{Statement-opinion}} by no-context model (though with less confidence) but the context-based model could recognize it as a \textit{\textbf{Yes-No-Question}} in the given dialogue. 
For the utterances \textit{utt4} and \textit{utt6}, the context-based model could correctly recognize the \textit{\textbf{Negative Answer}} and \textit{\textbf{Agree/Accept}} dialogue acts respectively but with very low confidence values. 
This shows a challenge to improve the context-based model.

\subsection{Key pointers}

The Web Interface is provided with different features, such as detailed analysis (we provide detailed output results of the models, at least the top 3 predictions with higher confidences).
Also, some examples are given to understand what one should expect from this demonstration.
The demo can be used for single utterance also, where the output will be produced only from the no-context model.
We do not record any data, as the application is for demonstration purpose only.

\section{Future work}

The demonstration of dialogue act recognition can be improved in several ways. 
First, we can combine several dialogue act corpora to improve the performance of the model. 
It would also be useful to use more performance but computationally expensive models as comparison like for \citet{bothe2018conversational,BOTHE18_525} or \citet{lee2016sequential} which will be added in future work.
This tool can be used for annotation purposes, given that the link for the particular task can be generated to be used as a standalone application.
Adding attention mechanisms to the models can also provide an additional feature to judge the importance of words and perhaps also of the utterances in the output results. 
Also adding an automatic speech recognition system would help to achieve on-the-go speech input for spoken utterances.

\section{Conclusions}

We present a first live demo for dialogue act recognition and discourse analysis of our approach. 
Our neural models surpass the baseline and some of the state-of-the-art results and we plan to improve them further in future research.
The live demo can be used for online conversational analysis, where classifying a single utterance or feeding a set of utterances from a conversation is possible with each alternating line corresponding to one speaker. 

Our goal was not only to present a novel live demo but also to develop an easy-to-use interface for conversational analysis and to provide a knowledge transfer tool. 
The modular architecture allows to integrate multiple interactive models trained on multiple/different corpora using separate Model Servers.
The Web Server can be extended to communicate with any number of Model Servers. 
Hence modular development and easy deployment of the systems is possible with such a simple architecture.





\section*{Acknowledgments}

This project has received funding from the European Union's Horizon 2020 research and innovation programme under the Marie Sklodowska-Curie grant agreement No 642667 (SECURE).  
\balance

\bibliography{emnlp2018}

\begin{thebibliography}{32}
\expandafter\ifx\csname natexlab\endcsname\relax\def\natexlab#1{#1}\fi

\bibitem[{Abadi et~al.(2016)Abadi, Barham, Chen, Chen, Davis, Dean, Devin,
  Ghemawat, Irving, Isard et~al.}]{abadi2016tensorflow}
Mart{\'\i}n Abadi, Paul Barham, Jianmin Chen, Zhifeng Chen, Andy Davis, Jeffrey
  Dean, Matthieu Devin, Sanjay Ghemawat, Geoffrey Irving, Michael Isard, et~al.
  2016.
\newblock {TensorFlow: A System for Large-Scale Machine Learning.}
\newblock In \emph{the Proc. of the 12th USENIX Symposium on Operating Systems
  Design and Implementation}, volume~16, pages 265--283.

\bibitem[{Allen and Core(1997)}]{allenCore1997}
James Allen and Mark Core. 1997.
\newblock {Draft of DAMSL: Dialogue Act Markup in Several Layers}.

\bibitem[{Austin(1962)}]{austin1962things}
John~L. Austin. 1962.
\newblock \emph{How to Do Things with Words}.
\newblock Oxford University Press.

\bibitem[{Bothe et~al.(2018{\natexlab{a}})Bothe, Magg, Weber, and
  Wermter}]{bothe2018conversational}
Chandrakant Bothe, Sven Magg, Cornelius Weber, and Stefan Wermter.
  2018{\natexlab{a}}.
\newblock {Conversational Analysis using Utterance-level Attention-based
  Bidirectional Recurrent Neural Networks}.
\newblock In \emph{Proc. of the Interspeech Conference}. ISCA.

\bibitem[{Bothe et~al.(2018{\natexlab{b}})Bothe, Weber, Magg, and
  Wermter}]{BOTHE18_525}
Chandrakant Bothe, Cornelius Weber, Sven Magg, and Stefan Wermter.
  2018{\natexlab{b}}.
\newblock {A Context-based Approach for Dialogue Act Recognition using Simple
  Recurrent Neural Networks}.
\newblock In \emph{Proc. of the Eleventh International Conference on Language
  Resources and Evaluation (LREC 2018)}, pages 1952--1957. ERLA.

\bibitem[{Chollet(2015)}]{Chollet2015Keras}
Fran{\c{c}}ois Chollet. 2015.
\newblock {Keras}.
\newblock In \emph{GitHub Repository}. GitHub.

\bibitem[{Chung et~al.(2017)Chung, Ahn, and Bengio}]{chung2017hierarchical}
Junyoung Chung, Sungjin Ahn, and Yoshua Bengio. 2017.
\newblock {Hierarchical Multiscale Recurrent Neural Networks}.
\newblock In \emph{Proc. of the International Conference on Learning
  Representations (ICLR 2017)}.

\bibitem[{El~Hihi and Bengio(1996)}]{el1996hierarchical}
Salah El~Hihi and Yoshua Bengio. 1996.
\newblock {Hierarchical Recurrent Neural Networks for Long-Term Dependencies}.
\newblock In \emph{Advances in Neural Information Processing Systems}, pages
  493--499.

\bibitem[{Elman(1990)}]{elman1990finding}
Jeffrey~L Elman. 1990.
\newblock Finding structure in time.
\newblock \emph{Cognitive Science}, 14(2):179--211.

\bibitem[{Godfrey et~al.(1992)Godfrey, Holliman, and
  McDaniel}]{godfrey1992switchboard}
John~J Godfrey, Edward~C Holliman, and Jane McDaniel. 1992.
\newblock {SWITCHBOARD: Telephone speech corpus for research and development}.
\newblock In \emph{Proc. of the International Conference on Acoustics, Speech,
  and Signal Processing}, volume~1, pages 517--520.

\bibitem[{Grau et~al.(2004)Grau, Sanchis, Castro, and Vilar}]{grau2004dialogue}
Sergio Grau, Emilio Sanchis, Maria~Jose Castro, and David Vilar. 2004.
\newblock {Dialogue act classification using a Bayesian approach}.
\newblock In \emph{Proc. of the 9th Conference Speech and Computer (SPECOM)}.

\bibitem[{Grinberg(2018)}]{grinberg2018flask}
Miguel Grinberg. 2018.
\newblock \emph{{Flask Web Development: Developing Web Applications with
  Python}}.
\newblock O'Reilly Media, Inc.

\bibitem[{Grosz(1982)}]{grosz1982discourse}
Barbara~J Grosz. 1982.
\newblock \emph{{Discourse Analysis (Chapter 5) Sublanguage: Studies of
  Language in Restricted Semantic Domains}}.
\newblock Walter de Gruyter, Berlin.

\bibitem[{Hochreiter and Schmidhuber(1997)}]{Hochreiter1997LongMemory}
Sepp Hochreiter and J{\"u}rgen Schmidhuber. 1997.
\newblock {Long Short-Term Memory}.
\newblock \emph{Neural Computation}, 9(8):1735--80.

\bibitem[{Holovaty and Kaplan-Moss(2009)}]{holovaty2009definitive}
Adrian Holovaty and Jacob Kaplan-Moss. 2009.
\newblock \emph{{The Definitive Guide to Django: Web Development Done Right}}.
\newblock Apress.

\bibitem[{Jurafsky(1997)}]{danjur1997swbddamsl}
Daniel Jurafsky. 1997.
\newblock {Switchboard SWBD-DAMSL Shallow-Discourse-Function Annotation Coders
  Manual, draft 13}.
\newblock \emph{Technical Report 97-01, University of Colorado Institute of
  Cognitive Science}, pages 225--233.

\bibitem[{Jurafsky et~al.(1998)Jurafsky, Shribergy, Fox, and
  Curl}]{Jurafsky1998}
Daniel Jurafsky, Elizabeth Shribergy, Barbara Fox, and Traci Curl. 1998.
\newblock {Lexical, Prosodic, and Syntactic Cues for Dialog Acts}.
\newblock In \emph{ACL/COLING Workshop on Discourse Relations and Discourse
  Markers}.

\bibitem[{Kalchbrenner and Blunsom(2013)}]{kalchbrenner2013recurrent}
Nal Kalchbrenner and Phil Blunsom. 2013.
\newblock {Recurrent Convolutional Neural Networks for Discourse
  Compositionality}.
\newblock In \emph{Proc. of the Workshop on Continuous Vector Space Models and
  their Compositionality, ACL}, pages 119--126.

\bibitem[{Kumar et~al.(2017)Kumar, Agarwal, Dasgupta, Joshi, and
  Kumar}]{kumar2017dasl}
Harshit Kumar, Arvind Agarwal, Riddhiman Dasgupta, Sachindra Joshi, and Arun
  Kumar. 2017.
\newblock {Dialogue Act Sequence Labeling using Hierarchical encoder with CRF}.
\newblock \emph{arXiv:1709.04250v2}.

\bibitem[{Kutuzov and Kuzmenko(2017)}]{kutuzov2017building}
Andrey Kutuzov and Elizaveta Kuzmenko. 2017.
\newblock {Building Web-Interfaces for Vector Semantic Models with the
  WebVectors Toolkit}.
\newblock In \emph{Proc. of the Software Demonstrations of the 15th Conference
  of the European Chapter of the Association for Computational Linguistics},
  pages 99--103.

\bibitem[{Lee and Dernoncourt(2016)}]{lee2016sequential}
Ji~Young Lee and Franck Dernoncourt. 2016.
\newblock {Sequential Short-Text Classification with Recurrent and
  Convolutional Neural Networks}.
\newblock \emph{arXiv:1603.03827}.

\bibitem[{Liu et~al.(2016)Liu, Qiu, and Huang}]{lai2015recurrent}
Pengfei Liu, Xipeng Qiu, and Xuanjing Huang. 2016.
\newblock {Recurrent Convolutional Neural Networks for Text Classificationn
  with Multi-Task Learning}.
\newblock In \emph{Proc. of the International Joint Conference on Artificial
  Intelligence}, pages 2873--2879.

\bibitem[{Liu et~al.(2017)Liu, Han, Tan, and Lei}]{ltAl2017EMNLP}
Yang Liu, Kun Han, Zhao Tan, and Yun Lei. 2017.
\newblock {Using Context Information for Dialog Act Classification in DNN
  Framework}.
\newblock In \emph{Proc. of the Conference on EMNLP}, pages 2160--2168. ACL.

\bibitem[{Loper and Bird(2002)}]{Loper2002NLTK:Toolkit}
Edward Loper and Steven Bird. 2002.
\newblock {NLTK: the Natural Language Toolkit}.
\newblock \emph{Proc. of the ACL-2 Workshop on Effective Tools and
  Methodologies for Teaching Natural Language Processing and Computational
  Linguistics}, 1:63--70.

\bibitem[{Manning et~al.(2014)Manning, Surdeanu, Bauer, Finkel, Bethard, and
  McClosky}]{manning2014stanford}
Christopher Manning, Mihai Surdeanu, John Bauer, Jenny Finkel, Steven Bethard,
  and David McClosky. 2014.
\newblock {The Stanford CoreNLP natural language processing toolkit}.
\newblock In \emph{Proc. of 52nd Annual Meeting of the Association for
  Computational Linguistics: System Demonstrations}, pages 55--60.

\bibitem[{Meng et~al.(2017)Meng, Mou, and Jin}]{Meng2017}
Zhao Meng, Lili Mou, and Zhi Jin. 2017.
\newblock {Hierarchical RNN with Static Sentence-Level Attention for Text-Based
  Speaker Change Detection}.
\newblock In \emph{Proc. of the ACM Conference on Information and Knowledge
  Management}, pages 2203--2206.

\bibitem[{Ortega and Vu(2017)}]{ortega2017neural}
Daniel Ortega and Ngoc~Thang Vu. 2017.
\newblock {Neural-based Context Representation Learning for Dialog Act
  Classification}.
\newblock In \emph{Proc. of the Conference of the Special Interest Group on
  Discourse and Dialogue}, pages 247--252.

\bibitem[{Sbis{\`a}(2002)}]{sbisa2002speech}
Marina Sbis{\`a}. 2002.
\newblock Speech acts in context.
\newblock \emph{Language \& Communication}, 22(4):421--436.

\bibitem[{Searle(1979)}]{searle1979}
John~R. Searle. 1979.
\newblock \emph{{Expression and Meaning: Studies in the Theory of Speech
  Acts}}.
\newblock Cambridge University Press.

\bibitem[{Socher et~al.(2013)Socher, Perelygin, Wu, Chuang, Manning, Ng, and
  Potts}]{socher2013recursive}
Richard Socher, Alex Perelygin, Jean Wu, Jason Chuang, Christopher~D Manning,
  Andrew Ng, and Christopher Potts. 2013.
\newblock {Recursive Deep Models for Semantic Compositionality Over a Sentiment
  Treebank}.
\newblock In \emph{Proc. of the Conference on Empirical Methods in Natural
  Language Processing}, pages 1631--1642.

\bibitem[{Stolcke et~al.(2000)Stolcke, Ries, Coccaro, Shriberg, Bates,
  Jurafsky, Taylor, Martin, Van Ess-Dykema, and Meteer}]{stolcke2000dialogue}
Andreas Stolcke, Klaus Ries, Noah Coccaro, Elizabeth Shriberg, Rebecca Bates,
  Daniel Jurafsky, Paul Taylor, Rachel Martin, Carol Van Ess-Dykema, and Marie
  Meteer. 2000.
\newblock {Dialogue Act Modeling for Automatic Tagging and Recognition of
  Conversational Speech}.
\newblock \emph{Computational Linguistics}, 26(3):339--373.

\bibitem[{Wermter(1995)}]{wermter1995hybrid}
Stefan Wermter. 1995.
\newblock \emph{{Hybrid Connectionist Natural Language Processing}}, volume~7.
\newblock Chapman \& Hall, London.

\end{thebibliography}
\bibliographystyle{acl_natbib_nourl}
\end{document}